\documentclass[10pt, a4paper]{article}

\usepackage[]{lrec-coling2024} 

\usepackage{csquotes}
\usepackage{makecell}

\usepackage{amsmath} 

\usepackage{hyperref}

\usepackage{dingbat}

\usepackage[multiple]{footmisc}
\usepackage{bigfoot}

\usepackage{comment}
\usepackage{graphicx}

\usepackage{booktabs}
\usepackage{multirow}
\usepackage[normalem]{ulem}
\useunder{\uline}{\ul}{}

\usepackage{xcolor,pifont}

\definecolor{brickred}{rgb}{0.745,0.176,0.176}
\definecolor{lightgreen}{rgb}{0.6, 0.9, 0.6}
\definecolor{mediumgreen}{rgb}{0.2, 0.7, 0.2}
\definecolor{darkgreen}{rgb}{0, 0.5, 0}

\newcommand{\ie}{\textit{i.e.}}
\newcommand{\eg}{\textit{e.g.}}

\usepackage[skins]{tcolorbox}
\usepackage{booktabs}

\title{Can Small Language Models Help Large Language Models Reason Better?: \textit{LM-Guided Chain-of-Thought}}

\name{\parbox{\linewidth}{\centering
Jooyoung Lee$^{1{\dagger}}$\thanks{$^\dagger$This work was done during the internship at Amazon.}, Fan Yang$^2$, Thanh Tran$^2$, Qian Hu$^2$ \\ 
Emre Barut$^2$, Kai-Wei Chang$^2$, Chengwei Su$^2$ }}

\address{Penn State University, PA, USA$^1$ \\
        Amazon AGI, MA, USA$^2$ \\
         jfl5838@psu.edu,
         \{fyaamz, tdt, huqia, ebarut, kaiwec, chengwes\}@amazon.com\\}

\abstract{
We introduce a novel framework, \textbf{LM-Guided CoT}, that leverages a lightweight (\ie, <1B) language model (LM) for guiding a black-box large (\ie, >10B) LM in reasoning tasks. Specifically, the lightweight LM first generates a rationale for each input instance. The Frozen large LM is then prompted to predict a task output based on the rationale generated by the lightweight LM. Our approach is resource-efficient in the sense that it only requires training the lightweight LM. We optimize the model through 1) knowledge distillation and 2) reinforcement learning from rationale-oriented and task-oriented reward signals. We assess our method with multi-hop extractive question answering (QA) benchmarks, HotpotQA, and 2WikiMultiHopQA. Experimental results show that our approach outperforms all baselines regarding answer prediction accuracy. We also find that reinforcement learning helps the model to produce higher-quality rationales with improved QA performance. 
\\ \newline  }

\begin{document}

\maketitleabstract

\newtcolorbox{mybox}{
  enhanced,
  colback=lightgray!15, 
  colframe=lightgray!15!, 
  fonttitle=\bfseries,
  colbacktitle=lightgray!85!black, 
  coltitle=white,
  attach boxed title to top left={yshift=-2mm, xshift=3mm},
  boxed title style={sharp corners},
  arc=1mm,
  drop fuzzy shadow
}

\section{Introduction}

Chain-of-Thought (CoT) prompting \cite{wei2022chain} has gained attention as a means to elicit the inherent reasoning abilities of a language model (LM). By prompting the models with \textit{``Let's think step by step''} following the actual task description, the model first produces intermediate reasoning steps and then predicts a task output. It has been shown to enhance the downstream task performance in complex reasoning domains such as arithmetic \cite{lewkowycz2022solving}, commonsense \cite{jung2022maieutic}, and symbolic \cite{khot2022decomposed} reasoning. However, there are several limitations to conventional CoT prompting. Firstly, the performance gains when compared to standard prompting (\ie, without \textit{``Let's think step by step''}) are only likely to emerge in very large LMs, preferably 100+ billion parameters \cite{wei2022emergent}. Moreover, the models may still generate low-quality rationales that are repetitive and vacuous \cite{ye2022unreliability}. This can be attributed to their tendencies to lack faithfulness to an input instance \cite{lanham2023measuring} and to produce unaligned rationales and answers \cite{wang2022self, turpin2023language}. 



\begin{figure}[t]
\centering
\includegraphics[width=\linewidth]{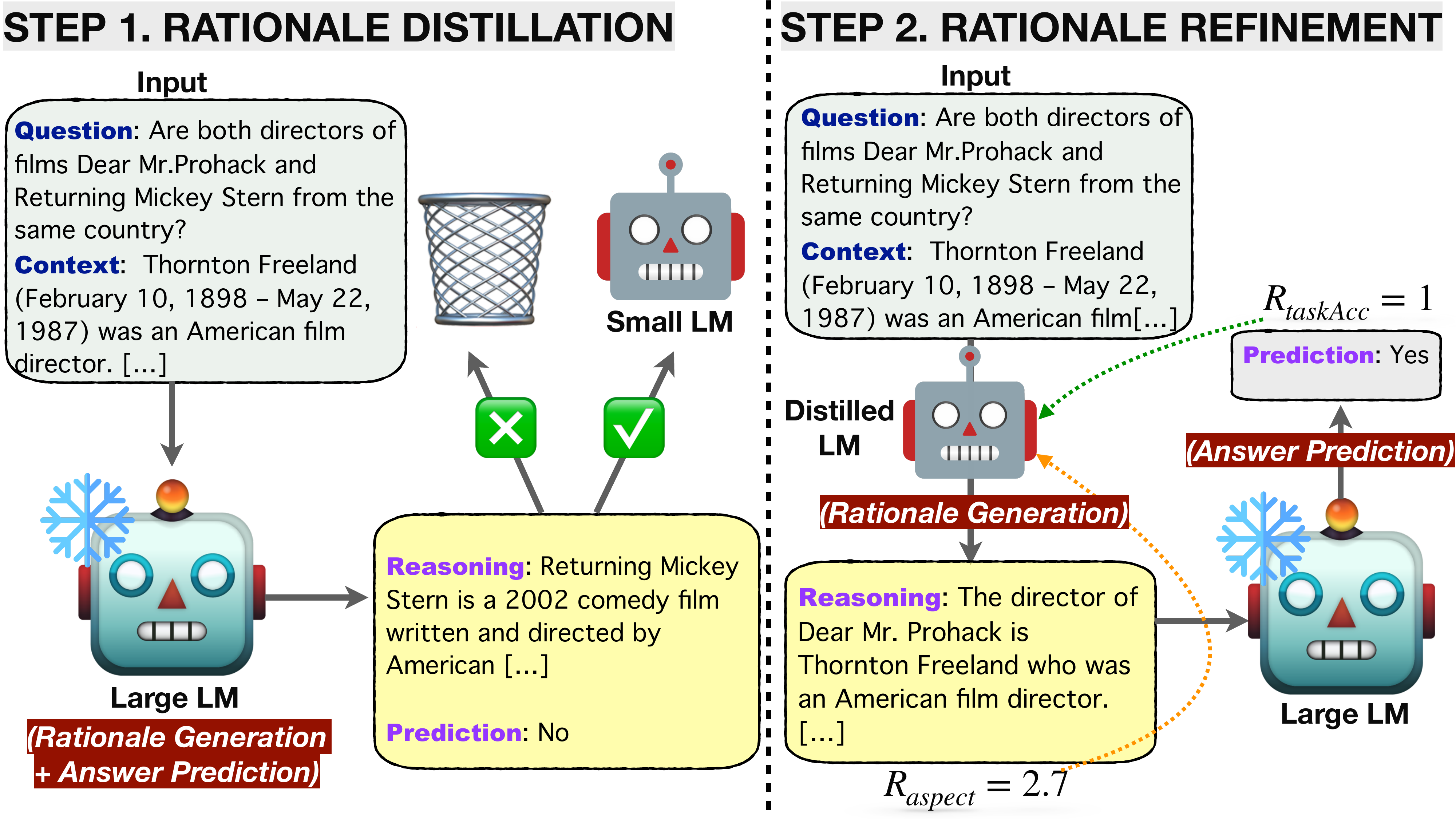}
\caption{Illustration of our proposed method.}
\vspace{-5mm}
\label{fig:pipeline}
\end{figure}

Given that CoT prompting is primarily effective with large LMs, rectifying these undesirable behaviors through direct model modifications is non-trivial, particularly with constrained computational resources. Hence, we propose \textbf{LM-guided CoT}, a novel framework that leverages two independent LMs (\ie, a small LM for rationale generation and a large LM for answer prediction) for CoT reasoning. 
As shown in Figure \ref{fig:pipeline}, we first employ a vanilla knowledge distillation (KD) technique to the small LM with rationales generated by the large LM (\S \ref{sec:rationale_distill}). 
This initial step helps narrow the gap in reasoning capabilities between the smaller and larger LMs to a certain extent. To further improve the quality of rationales generated by the knowledge-distilled LM, we establish fine-grained measurements concerning 8 rationale-specific aspects (relevance, actuality, logicality, consistency, coherence, fluency, naturalness, and readability) and use them to optimize the knowledge-distilled LM with reinforcement learning (RL) (\S \ref{sec:rationale_refinement}). 

We conduct experiments on an extractive multi-hop question answering (QA) task using two popular benchmarks, HotpotQA \citelanguageresource{yang2018hotpotqa} and 2WikiMultiHopQA \citelanguageresource{ho2020constructing}. Although our framework can be flexibly applied to a wide range of LMs, we use FLAN-T5 \cite{longpre2023flan} models in this work because they are open-source and instruction-tuned on both QA and CoT data, enabling us to use it off-the-shelf without additional training or prompt engineering.\footnote{We also ran a small experiment with the \href{https://huggingface.co/VMware/open-llama-7b-open-instruct}{instruction-tuned Open LLaMa (7B) model} and confirmed that its performance is significantly worse than FLAN-T5 in a zero-shot setting.} Our experiment results show that \uline{LM-guided CoT prompting outperforms both the standard prompting and the original CoT prompting.} More precisely, we find that (1) LM-guided CoT with KD and self-consistency (SC) decoding strategy \cite{wang2022self} maximizes the performance gain; (2) RL contributes to a slight increase in overall rationale quality and task performance; (3) choosing the highest-quality rationales for the large LM does not always guarantee improved task performance. This work presents a unique alternative to the direct optimization of the large LM through fine-tuning the comparatively smaller LM. Moreover, the clear separation of two fundamental sub-tasks within CoT reasoning grants practitioners greater control over each task. 

\section{Related Work}
\textbf{Rationale Distillation.} For computation efficiency or task performance, recent literature has explored methods to improve small LMs' reasoning abilities. \citet{li2023symbolic, shridhar2023distilling, ma2023sci} have experimented with rationale distillation, where a small student LM learns from a large teacher LM to generate CoT rationales. While these studies have mainly concentrated on comparing its performance in downstream tasks against that of large LMs, there has been limited investigation into addressing errors in the generated rationales that might have been inherited from the teacher model.

\noindent \textbf{Rationale Evaluation and Refinement.} In contexts beyond rationale distillation, there have been growing efforts to unravel which aspects of the generated reasoning steps contribute to the downstream task performance. \citet{wang2022towards} report that rationales' logicality and relevance to the query are key factors in successful CoT reasoning. Few studies have measured the validity of reasoning steps from the lens of more diverse aspects like informativeness, coherence, and repetition, etc \cite{golovneva2022roscoe, prasad2023receval}. 
While RL has gained popularity as an approach for addressing misaligned behaviors in LMs, the field of rationale correction has seen limited research.

\section{LM-guided Chain-of-Thought}
The proposed framework consists of two LMs: a lightweight model $M^S$ that focuses on generating the optimal rationale given an input instance, and a black-box large model $M^{L}$ that predicts an output based on the rationale generated by $M^S$. 

\begin{table*}[]
\small
\centering
\begin{tabular}{|c|l|}
\hline
\textbf{Aspects}     & \textbf{Descriptions}                                                                                                                              \\ \hline
Factuality  & Percentage (0.0-1.0) measuring if the reasoning is grounded based on the context \textcolor{blue}{\textit{(Input: $c$ \& $r'$)}}\\ \hline
Relevance   & Percentage (0.0-1.0) measuring if the reasoning is relevant to the question \textcolor{blue}{\textit{(Input: $q$ \& $r'$)}}                                \\ \hline
Logicality  & Binary (0 or 1) measuring if the reasoning is logical and can reach a final answer   \\ \hline
Consistency & Binary (0 or 1) measuring if the reasoning remains consistent and coherent \textcolor{blue}{\textit{(Input:  $q$ \& $r'$)}}                           \\ \hline
Coherence   & Binary (0 or 1) measuring if the  reasoning is without redundant information \textcolor{blue}{\textit{(Input:  $q$ \& $r'$)}}  \\ \hline
Fluency     & Binary (0 or 1) measuring if the reasoning is well-written and grammatically correct \textcolor{blue}{\textit{(Input: $r'$)}}            \\ \hline
Naturalness & Binary (0 or 1) measuring if the reasoning is natural and human-like  \textcolor{blue}{\textit{(Input: $r'$)}}              \\ \hline
Readability & Binary (0 or 1) measuring if the  reasoning is easy to follow and understandable \textcolor{blue}{\textit{(Input: $r'$)}}           \\ \hline
\end{tabular}
\caption{Descriptions of 8 rationale aspects used for evaluation. $q$, $c$, and $r'$ denote a question, context, and a corresponding rationale generated by the small LM, respectively.}
\label{tab:metric_table}
\end{table*}

\subsection{Rationale Distillation} 
\label{sec:rationale_distill}

\textbf{Rationale Generation.} In general, multi-hop extractive QA datasets contain a list of questions $Q$, contexts $C$, and corresponding ground truth answers $A$. For each input ($q$, $c$)-output ($a$) pair, we need a corresponding ground truth rationale $r$ to train $M^S$ for rationale generation in a supervised manner. However, most QA benchmarks do not provide $r$. Given that manual annotation for $r$ is labor-intensive and time-consuming, we 
\begin{mybox}
\footnotesize
Based on the provided context, answer the following question (Q) by reasoning step-by-step. 
\newline \textbf{Context}: $c$
\newline \textbf{Q}: $q$
\newline \textbf{A} : Let’s think step by step.
\end{mybox}

\noindent For generation, we use greedy decoding; \ie, choosing the most plausible token at each generation step.

\noindent \textbf{Rationale Filtering and Training.} When it comes to knowledge distillation, data cleaning processes play a crucial role in preventing errors or noises included in the generation from the teacher LM being inherited to the student LM. Thus, we filter samples associated with unfaithful responses to the prompt (\ie, not providing rationales prior to providing a final answer) and inaccurate answer prediction. Finally, we instruction-tune $M^{S}$ using the following prompt:

\begin{mybox}
\footnotesize
Given a question (Q) and a context, generate a chain of reasoning step by step to answer the question. 
\newline \textbf{Context}: $c$
\newline \textbf{Q}: $q$
\newline \textbf{Reasoning}: $r'$
\end{mybox}

\noindent For the rest of the paper, we denote the rationale-distilled model as $M^*$.

\subsection{Rationale Refinement}
\label{sec:rationale_refinement}

\noindent \textbf{Annotation for Rationale Quality Measurement.}
Inspired by previous text and rationale generation evaluation metrics \cite{golovneva2022roscoe, fu2023gptscore}, we attempt to quantify 8 linguistic aspects (factuality, relevance, logicality, consistency, coherence, fluency, naturalness, readability) of rationales generated by $M^*$ in \S \ref{sec:rationale_distill}. Let us denote $r^*$ as $M^*$-generated reasoning. Since there is no ground truth rationale $r$ available for comparison, our metrics are reference-free, utilizing a pair of $r^*$ and one of existing inputs ($q$ or $c$). Table \ref{tab:metric_table} describes each aspect type and input combinations. As we intend to use these metrics for reward scoring in RL, it is critical to have an accurate metric. Hence, we obtain a small set (n=100) of gold labels for all aspect types through human annotation. A detailed description of the annotation process and results is reported in \S \ref{sec:appendix_human_annotation}.

\noindent \textbf{Automatic Measurement for Rationale Quality.}
Manual annotation offers substantial value, but within the context of RL, it becomes notably challenging due to frequent reward scoring. Therefore, we probe several ways to automate this process. \citet{ye2022unreliability} introduced the simple yet effective approach for assessing factuality and relevance through token-level lexical overlap. We follow their method for factuality and relevance measurement. For the remaining 6 categories, two methods are considered. The first approach is to harness a large LM to function as reference-free NLG evaluators, inspired by recent works (\eg, \citet{liu2023gpteval}, \citet{wang2023chatgpt}). The second approach, in contrast, involves training a simple machine learning classifier using human-annotated data. Both approaches are comprehensively described in \S \ref{sec:appendix_automatic_annotation}. Due to inference time efficiency and higher alignment scores to human annotators (see Table \ref{tab:comparison_results}), we resort to the second method for all our experiments.



\noindent \textbf{RL for Rationale Refinement.}
We next detail how to utilize established evaluation metrics as reward signals to update the knowledge-distilled $M^*$ with Proximal Policy Optimization (PPO) \cite{schulman2017proximal}. 
Given each input ($q$, $c$)-output ($a$) pair from training data, we first prompt $M^*$ to generate a corresponding rationale $r^*$. During the generation process, an aspect-specific reward (denoted as $R_{aspect}$) is measured by aggregating all values returned from automatic evaluation metrics.\footnote{We also test with normalization or weighted summation for the scores, but they did not affect the performance. }
We then pass $r^*$ to $M^L$ to retrieve the answer prediction $a^*$ and compute a task-specific reward (denoted as $R_{taskAcc}$). Specifically, we leverage the F1 score between the predicted answer and the ground truth answer: 
\begin{align*}
R_{\text{taskAcc}} = 
\begin{cases}
    1 & \text{if } F1(a, a^*) > 0.5, \\
    0 & \text{else}.
\end{cases}
\end{align*}
A final reward score for model training is the summation of $R_{aspect}$ and $R_{taskAcc}$. 
Following \citet{stiennon2020learning}, we also incorporate penalties based on the Kullback Leibler (KL) divergence between the learned policy LM and $M^S$.


\begin{table*}[]
\small
\centering
\begin{tabular}{@{}ccccclccc@{}}
\toprule
\multirow{2}{*}{Prompt}                                                                          & \multirow{2}{*}{{\makecell{Rationale\\Provision?}} } & \multicolumn{3}{c}{HotpotQA}                                                                  &  & \multicolumn{3}{c}{2WikiMultiHopQA}                                                                    \\ \cmidrule(lr){3-5} \cmidrule(l){7-9} 
                                                                                                 &                             & EM             & F1             & \begin{tabular}[c]{@{}c@{}}Answer \\ Inclusion\end{tabular} &  & EM                      & F1             & \begin{tabular}[c]{@{}c@{}}Answer \\ Inclusion\end{tabular} \\ \midrule
standard prompting                                                                               & \textcolor{red}{\ding{55}}                           & 0.5            & \textbf{0.714} & 0.583                                                       &  & 0.5                     & 0.625          & 0.647                                                       \\
CoT prompting                                                                                    & \textcolor{darkgreen}{\ding{51}}                            & 0.483          & 0.686          & 0.611                                                       &  & 0.4                     & 0.532          & 0.561                                                       \\
CoT prompting + SC                                                                & \textcolor{red}{\ding{55}}                           & 0.503          & 0.70           & 0.624                                                       &  & 0.471                   & 0.603          & 0.625                                                       \\ \midrule
LM-guided CoT prompting (KD)                                                                    & \textcolor{darkgreen}{\ding{51}}                            & 0.507          & 0.702          & 0.625                                                       &  & 0.506                   & 0.626          & 0.661                                                       \\
LM-guided CoT prompting (KD + SC)     & \textcolor{red}{\ding{55}}                           & \textbf{0.513} & \textbf{0.714} & \textbf{0.635}                                              &  & \textbf{0.524} & \textbf{0.644} & \textbf{0.679}                                              \\
\begin{tabular}[c]{@{}c@{}}LM-guided CoT prompting \\ (KD + $R_{aspect}$)\end{tabular}             & \textcolor{darkgreen}{\ding{51}}                           & 0.503          & 0.698          & 0.625                                                       &  & {\ul 0.507}             & {\ul 0.631}    & {\ul 0.665}                                                 \\
\begin{tabular}[c]{@{}c@{}}LM-guided CoT prompting\\  (KD + $R_{aspect}$ + $R_{taskAcc}$)\end{tabular} & \textcolor{darkgreen}{\ding{51}}                            & {\ul 0.508}    & {\ul 0.704}    & {\ul 0.627}                                                 &  & 0.503                   & 0.622          & 0.657                                                       \\
\begin{tabular}[c]{@{}c@{}}LM-guided CoT prompting\\ (KD +$R_{aspect}$ + ranking)\end{tabular}     & \textcolor{darkgreen}{\ding{51}}                            & 0.5            & 0.698          & 0.623                                                       &  & 0.501                   & 0.619          & 0.653                                                       \\ \bottomrule
\end{tabular}
\caption{Answer prediction performance results of baselines and our approach. We regard SC decoding as a non-rationale provision because this method can result in multiple variations of rationales, rather than a single one. Values in bold represent the highest scores and underlined values are the second highest scores.}
\label{tab:results}
\end{table*}

\begin{table*}[]
\small
\begin{tabular}{|c|c|l|}
\hline
\textbf{Type}                                                                                          & \textbf{Description}                                                                                                                     & \multicolumn{1}{c|}{\textbf{Template}}                                                                                                                                                                                                                                      \\ \hline
\begin{tabular}[c]{@{}c@{}}Standard \\ prompting\end{tabular}                                           & \begin{tabular}[c]{@{}c@{}}Directly predicting the answer\\  based on input\end{tabular}                                        & \begin{tabular}[c]{@{}l@{}}Based on the provided context, answer the following question (Q). \\ Context: \color{darkgreen}{\boldmath$c$} \\ Q: \color{magenta}\boldmath{$q$}\\ A:\end{tabular}                                                                                                          \\ \hline
\begin{tabular}[c]{@{}c@{}}CoT \\ prompting\end{tabular}                                                & \begin{tabular}[c]{@{}c@{}}Predicting the answer after \\ generating the reasoning\end{tabular}                                 & \begin{tabular}[c]{@{}l@{}}Based on the provided context, answer the following question (Q)\\ by reasoning step-by-step. \\ Context: \color{darkgreen}{\boldmath$c$} \\ Q: \color{magenta}{\boldmath$q$}\\ A : Let’s think step by step.\end{tabular}                                                   \\ \hline
{\begin{tabular}[c]{@{}c@{}}LM-guided \\ CoT prompting \\ \color{brickred}{\textbf{\textit{(our method)}}}\end{tabular}} & \begin{tabular}[c]{@{}c@{}}Predicting the answer with \\ conditional generation upon \\ the LM-generated reasoning\end{tabular} & \begin{tabular}[c]{@{}l@{}}Based on the provided context, answer the following question (Q) \\ by reasoning step-by-step. \\ Context: \color{darkgreen}{\boldmath$c$} \\ Q: \color{magenta}{\boldmath$q$}\\ A : Let’s think step by step. \color{orange}{\boldmath$r'$} \color{black}{. Hence, the answer is}\end{tabular} \\ \hline
\end{tabular}
\caption{Descriptions and templates of each prompt used for the answer prediction task. \color{magenta}{\boldmath{$q$}}\color{black}{,} \color{darkgreen}{\boldmath{$c$}}\color{black}{, and} \color{orange}\boldmath{$r'$} \color{black}{denote a question, context, and a corresponding rationale generated by the small LM, respectively.}}
\label{tab:prompt_template}
\end{table*}

\section{Experiments and Results}
\subsection{Experimental Setup}
\textbf{Model and Dataset.} we utilize FLAN-T5 small (80M) for $M^S$ and FLAN-T5 XXL (11B) for $M^L$.
Both HotpotQA \cite{yang2018hotpotqa} and 2WikiMultiHopQA \cite{ho2020constructing} consist of an input question, and an answer, along with 9-10 context paragraphs with supportiveness labels indicating whether the paragraph contains supporting facts. Due to the input token size limitation of FLAN-T5, we only use supporting paragraphs as context. 

\noindent \textbf{Training \& Evaluation Setup.} We use a randomly sampled subset of training data from two datasets (15K samples per data) for model training. After filtering unqualified reasoning, it results in 23K samples. All training-related hyperparameters can be found in \S \ref{sec:appendix_training_config}. According to our preliminary experiments, the impact of CoT prompting appeared to be diminished due to two potential factors: (1) questions being overly simplistic, obscuring the significance of intermediate reasoning processes; (2) LMs already possessing pertinent background information (\eg, Flan-T5 is fine-tuned on various question-answering datasets). To prevent models from answering based on parametric memory, we attempt to make the existing evaluation data more challenging, similar to the approaches taken by \citet{ye2022unreliability} and \citet{zhao2023verify}. For each dataset, we leverage the prediction outcomes of standard prompting and select 1000 input instances that $M^L$ answered correctly and an additional 1000 from questions where the model provided incorrect responses. This results in a total of 2000 samples for evaluation.



\noindent \textbf{Baselines and Evaluation Metrics.}  We use standard prompting and CoT prompting as our baselines (see Table \ref{tab:prompt_template}). We also experiment with the SC decoding strategy \cite{wang2022self}, which samples multiple reasoning paths (n=10) and selects the most consistent answer. 
For evaluation, we report three metrics for the answer prediction task: (1) exact match (EM), computing whether the prediction exactly matches the ground truth answer, (2) F1, computing the average word overlap between the prediction and ground truth answer, and (3) answer inclusion\footnote{We included this measurement because models often provide more extensive responses (\eg, ground truth: World's Best Goalkeeper $\rightarrow$ generation: IFFHS World's Best Goalkeeper). This may be explained by our usage of contextual paragraphs as input.}, computing whether the ground truth answer is mentioned in the prediction. 

\subsection{Results}


\textbf{Baseline Performance.}
As shown in Table \ref{tab:results}, we find that $M^L$ (equivalent to FLAN-T5 XXL) does not benefit from the original CoT prompting, as its EM and F1 scores dropped in both datasets (except for answer inclusion score for HotpotQA) when compared to standard prompting. This is consistent with previous research findings that models with less than 50B parameters exhibit limited reasoning capabilities. We also observe that the performance drop is more significant with 2WikiMultihopQA (nearly 10\% for EM and F1) than HotpotQA. Based on our manual inspection of incorrect predictions, $M^L$ was prone to repeat sentences in the context and fail to provide a final answer to the questions. This hints that, when context gets too long, models face difficulties in digesting the content and establishing valid reasoning steps. Overall, SC CoT prompting enables the model to noticeably recover from answer prediction errors, especially for 2WikiMultihopQA.

\begin{figure}[h]
\centering
\includegraphics[width=\linewidth]{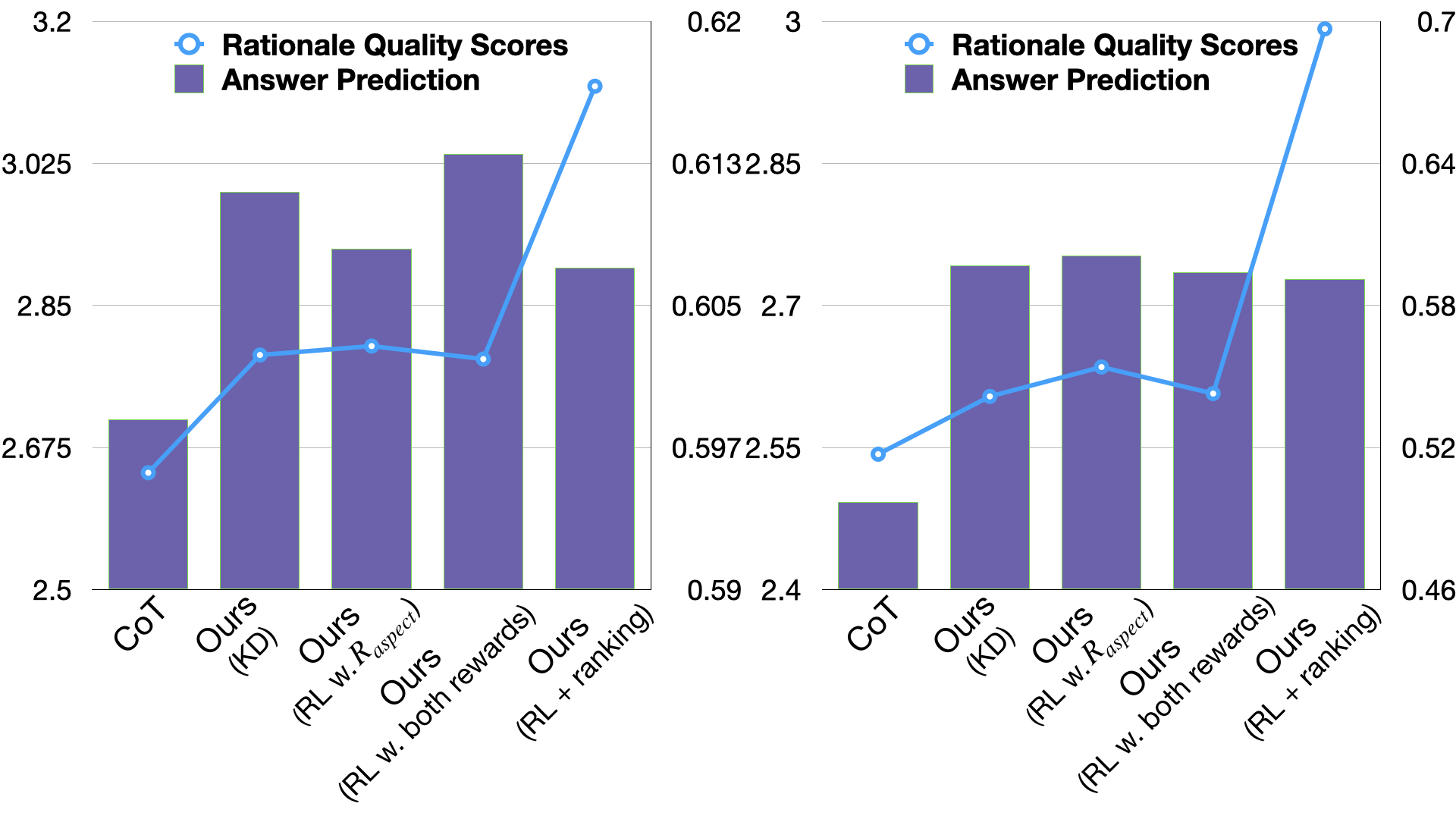}
\caption{Average answer prediction performance (across three evaluation metrics) and average rationale quality scores (\ie, $R_{aspect}$) for HotpotQA (left) and 2WikiMultiHopQA (right). The right y-axis represents the mean answer prediction scores, and the left y-axis represents the mean rationale quality scores.}
\label{fig:performance_reward}
\end{figure}

\noindent \textbf{LM-guided CoT Performance.}
\noindent Table \ref{tab:results} shows a comprehensive breakdown of our method's performance. Additionally, we explore an extension of our approach, which involves sampling multiple reasoning paths and subsequently ranking the most optimal rationales. Our method with only KD outperforms the original CoT prompting with 2\% gain for HotpotQA and 10\% for 2WikiMultiHopQA, respectively. Figure \ref{fig:performance_reward} illustrates the respective rationale qualities of all prompting techniques, reinforcing the effectiveness of our method in enhancing both answer prediction and rationale qualities. When employing the original CoT prompting for questions with lengthy contexts, models frequently recycle sentences from the provided context and struggle to deliver a conclusive answer to the question. This trend is mitigated by our approach, resulting in a significant decrease in error rates. It also surpasses the performance of CoT prompting + SC and is on par with standard prompting in terms of EM and F1. For the answer inclusion score, LM-guided CoT prompting is slightly higher (1-2\%) than standard prompting. Furthermore, LM-guided CoT prompting + SC achieves the highest performance across all settings.

\noindent As shown in Figure \ref{fig:performance_reward}, the implementation of RL enables the model to achieve additional improvements in both rationale qualities and task performance. 
However, in line with \citet{joshi2023machine}’s findings, a slight decrease in task performance is observed at the cost of maximized rationale qualities when selecting top-quality rationales. There may be several underlying factors involved (\eg, models’ unfaithfulness), but this is not the scope of this work.


\section{Conclusion}
\textbf{LM-Guided CoT} is a novel framework that decomposes a conventional CoT prompting into two steps using two models: (1) rationale generation and (2) answer prediction. This includes distilling the reasoning ability from a large LM to a small LM and further optimizing it with RL. The results reveal that our method outperforms all baselines, highlighting its potential to serve as an effective and resource-efficient approach to tackle challenges within the CoT prompting paradigm. Meanwhile, we also find that selecting top-quality rationales for answer prediction may not consistently boost task performance. This prompts the need to explore a more harmonious balance between LM-generated rationale utilities and overall task performance.

\section{Limitations}
Although our framework can seamlessly accommodate various model combinations in a plug-and-play manner, we have restricted our experimental reporting to FLAN-T5. In a similar vein, this work only explores the task of multi-hop QA, leaving an open question about generalizability to other reasoning tasks. We anticipate future research endeavors to extend the application of our approach across diverse domains requiring sophisticated reasoning. Lastly, due to resource constraints, we were unable to collect extensive human annotations for established aspect evaluation metrics.


\section{Ethical Considerations}

All the datasets that we use in our work are publicly available, and we have given appropriate credit to the original authors throughout the paper. We acknowledge that occasionally, the generated rationales may include non-factual and offensive statements. As we do not plan on distributing these artifacts, it effectively reduces the potential for harm.

\section{Acknowledgments}
We thank all the reviewers for providing valuable feedback.

\section{Bibliographical References}\label{sec:reference}

\bibliographystyle{lrec-coling2024-natbib}
\bibliography{lrec-coling2024-example}

\begin{thebibliography}{2}
\expandafter\ifx\csname natexlab\endcsname\relax\def\natexlab#1{#1}\fi

\bibitem[{Ho et~al.(2020)Ho, Nguyen, Sugawara, and Aizawa}]{ho2020constructing}
Ho, Xanh and Nguyen, Anh-Khoa Duong and Sugawara, Saku and Aizawa, Akiko. 2020.
\newblock \emph{Constructing a multi-hop QA dataset for comprehensive evaluation of reasoning steps}.

\bibitem[{Yang et~al.(2018)Yang, Qi, Zhang, Bengio, Cohen, Salakhutdinov, and Manning}]{yang2018hotpotqa}
Yang, Zhilin and Qi, Peng and Zhang, Saizheng and Bengio, Yoshua and Cohen, William W and Salakhutdinov, Ruslan and Manning, Christopher D. 2018.
\newblock \emph{HotpotQA: A dataset for diverse, explainable multi-hop question answering}.

\end{thebibliography}


\begin{thebibliography}{24}
\expandafter\ifx\csname natexlab\endcsname\relax\def\natexlab#1{#1}\fi

\bibitem[{Fleiss(1971)}]{fleiss1971measuring}
Joseph~L Fleiss. 1971.
\newblock Measuring nominal scale agreement among many raters.
\newblock \emph{Psychological bulletin}, 76(5):378.

\bibitem[{Fu et~al.(2023)Fu, Ng, Jiang, and Liu}]{fu2023gptscore}
Jinlan Fu, See-Kiong Ng, Zhengbao Jiang, and Pengfei Liu. 2023.
\newblock Gptscore: Evaluate as you desire.
\newblock \emph{arXiv preprint arXiv:2302.04166}.

\bibitem[{Golovneva et~al.(2022)Golovneva, Chen, Poff, Corredor, Zettlemoyer, Fazel-Zarandi, and Celikyilmaz}]{golovneva2022roscoe}
Olga Golovneva, Moya Chen, Spencer Poff, Martin Corredor, Luke Zettlemoyer, Maryam Fazel-Zarandi, and Asli Celikyilmaz. 2022.
\newblock Roscoe: A suite of metrics for scoring step-by-step reasoning.
\newblock \emph{In The Eleventh International Conference on Learning Representations}.

\bibitem[{Joshi et~al.(2023)Joshi, Liu, Ramnath, Chan, Tong, Nie, Wang, Choi, and Ren}]{joshi2023machine}
Brihi Joshi, Ziyi Liu, Sahana Ramnath, Aaron Chan, Zhewei Tong, Shaoliang Nie, Qifan Wang, Yejin Choi, and Xiang Ren. 2023.
\newblock Are machine rationales (not) useful to humans? measuring and improving human utility of free-text rationales.
\newblock \emph{arXiv preprint arXiv:2305.07095}.

\bibitem[{Jung et~al.(2022)Jung, Qin, Welleck, Brahman, Bhagavatula, Bras, and Choi}]{jung2022maieutic}
Jaehun Jung, Lianhui Qin, Sean Welleck, Faeze Brahman, Chandra Bhagavatula, Ronan~Le Bras, and Yejin Choi. 2022.
\newblock Maieutic prompting: Logically consistent reasoning with recursive explanations.
\newblock \emph{arXiv preprint arXiv:2205.11822}.

\bibitem[{Khot et~al.(2022)Khot, Trivedi, Finlayson, Fu, Richardson, Clark, and Sabharwal}]{khot2022decomposed}
Tushar Khot, Harsh Trivedi, Matthew Finlayson, Yao Fu, Kyle Richardson, Peter Clark, and Ashish Sabharwal. 2022.
\newblock Decomposed prompting: A modular approach for solving complex tasks.
\newblock \emph{arXiv preprint arXiv:2210.02406}.

\bibitem[{Lanham et~al.(2023)Lanham, Chen, Radhakrishnan, Steiner, Denison, Hernandez, Li, Durmus, Hubinger, Kernion et~al.}]{lanham2023measuring}
Tamera Lanham, Anna Chen, Ansh Radhakrishnan, Benoit Steiner, Carson Denison, Danny Hernandez, Dustin Li, Esin Durmus, Evan Hubinger, Jackson Kernion, et~al. 2023.
\newblock Measuring faithfulness in chain-of-thought reasoning.
\newblock \emph{arXiv preprint arXiv:2307.13702}.

\bibitem[{Lewkowycz et~al.(2022)Lewkowycz, Andreassen, Dohan, Dyer, Michalewski, Ramasesh, Slone, Anil, Schlag, Gutman-Solo et~al.}]{lewkowycz2022solving}
Aitor Lewkowycz, Anders Andreassen, David Dohan, Ethan Dyer, Henryk Michalewski, Vinay Ramasesh, Ambrose Slone, Cem Anil, Imanol Schlag, Theo Gutman-Solo, et~al. 2022.
\newblock Solving quantitative reasoning problems with language models.
\newblock \emph{Advances in Neural Information Processing Systems}, 35:3843--3857.

\bibitem[{Li et~al.(2023)Li, Hessel, Yu, Ren, Chang, and Choi}]{li2023symbolic}
Liunian~Harold Li, Jack Hessel, Youngjae Yu, Xiang Ren, Kai-Wei Chang, and Yejin Choi. 2023.
\newblock Symbolic chain-of-thought distillation: Small models can also" think" step-by-step.
\newblock \emph{arXiv preprint arXiv:2306.14050}.

\bibitem[{Liu et~al.(2023)Liu, Iter, Xu, Wang, Xu, and Zhu}]{liu2023gpteval}
Yang Liu, Dan Iter, Yichong Xu, Shuohang Wang, Ruochen Xu, and Chenguang Zhu. 2023.
\newblock Gpteval: Nlg evaluation using gpt-4 with better human alignment.
\newblock \emph{arXiv preprint arXiv:2303.16634}.

\bibitem[{Longpre et~al.(2023)Longpre, Hou, Vu, Webson, Chung, Tay, Zhou, Le, Zoph, Wei et~al.}]{longpre2023flan}
Shayne Longpre, Le~Hou, Tu~Vu, Albert Webson, Hyung~Won Chung, Yi~Tay, Denny Zhou, Quoc~V Le, Barret Zoph, Jason Wei, et~al. 2023.
\newblock The flan collection: Designing data and methods for effective instruction tuning.
\newblock \emph{arXiv preprint arXiv:2301.13688}.

\bibitem[{Ma et~al.(2023)Ma, Jiang, and Fan}]{ma2023sci}
Yuhan Ma, Haiqi Jiang, and Chenyou Fan. 2023.
\newblock Sci-cot: Leveraging large language models for enhanced knowledge distillation in small models for scientific qa.
\newblock \emph{arXiv preprint arXiv:2308.04679}.

\bibitem[{Prasad et~al.(2023)Prasad, Saha, Zhou, and Bansal}]{prasad2023receval}
Archiki Prasad, Swarnadeep Saha, Xiang Zhou, and Mohit Bansal. 2023.
\newblock Receval: Evaluating reasoning chains via correctness and informativeness.
\newblock \emph{arXiv preprint arXiv:2304.10703}.

\bibitem[{Schulman et~al.(2017)Schulman, Wolski, Dhariwal, Radford, and Klimov}]{schulman2017proximal}
John Schulman, Filip Wolski, Prafulla Dhariwal, Alec Radford, and Oleg Klimov. 2017.
\newblock Proximal policy optimization algorithms.
\newblock \emph{arXiv preprint arXiv:1707.06347}.

\bibitem[{Shridhar et~al.(2023)Shridhar, Stolfo, and Sachan}]{shridhar2023distilling}
Kumar Shridhar, Alessandro Stolfo, and Mrinmaya Sachan. 2023.
\newblock Distilling reasoning capabilities into smaller language models.
\newblock In \emph{Findings of the Association for Computational Linguistics: ACL 2023}, pages 7059--7073.

\bibitem[{Stiennon et~al.(2020)Stiennon, Ouyang, Wu, Ziegler, Lowe, Voss, Radford, Amodei, and Christiano}]{stiennon2020learning}
Nisan Stiennon, Long Ouyang, Jeffrey Wu, Daniel Ziegler, Ryan Lowe, Chelsea Voss, Alec Radford, Dario Amodei, and Paul~F Christiano. 2020.
\newblock Learning to summarize with human feedback.
\newblock \emph{Advances in Neural Information Processing Systems}, 33:3008--3021.

\bibitem[{Turpin et~al.(2023)Turpin, Michael, Perez, and Bowman}]{turpin2023language}
Miles Turpin, Julian Michael, Ethan Perez, and Samuel~R Bowman. 2023.
\newblock Language models don't always say what they think: Unfaithful explanations in chain-of-thought prompting.
\newblock \emph{arXiv preprint arXiv:2305.04388}.

\bibitem[{Wang et~al.(2022{\natexlab{a}})Wang, Min, Deng, Shen, Wu, Zettlemoyer, and Sun}]{wang2022towards}
Boshi Wang, Sewon Min, Xiang Deng, Jiaming Shen, You Wu, Luke Zettlemoyer, and Huan Sun. 2022{\natexlab{a}}.
\newblock Towards understanding chain-of-thought prompting: An empirical study of what matters.
\newblock \emph{arXiv preprint arXiv:2212.10001}.

\bibitem[{Wang et~al.(2023)Wang, Liang, Meng, Shi, Li, Xu, Qu, and Zhou}]{wang2023chatgpt}
Jiaan Wang, Yunlong Liang, Fandong Meng, Haoxiang Shi, Zhixu Li, Jinan Xu, Jianfeng Qu, and Jie Zhou. 2023.
\newblock Is chatgpt a good nlg evaluator? a preliminary study.
\newblock \emph{arXiv preprint arXiv:2303.04048}.

\bibitem[{Wang et~al.(2022{\natexlab{b}})Wang, Wei, Schuurmans, Le, Chi, Narang, Chowdhery, and Zhou}]{wang2022self}
Xuezhi Wang, Jason Wei, Dale Schuurmans, Quoc Le, Ed~Chi, Sharan Narang, Aakanksha Chowdhery, and Denny Zhou. 2022{\natexlab{b}}.
\newblock Self-consistency improves chain of thought reasoning in language models.
\newblock \emph{arXiv preprint arXiv:2203.11171}.

\bibitem[{Wei et~al.(2022{\natexlab{a}})Wei, Tay, Bommasani, Raffel, Zoph, Borgeaud, Yogatama, Bosma, Zhou, Metzler et~al.}]{wei2022emergent}
Jason Wei, Yi~Tay, Rishi Bommasani, Colin Raffel, Barret Zoph, Sebastian Borgeaud, Dani Yogatama, Maarten Bosma, Denny Zhou, Donald Metzler, et~al. 2022{\natexlab{a}}.
\newblock Emergent abilities of large language models.
\newblock \emph{arXiv preprint arXiv:2206.07682}.

\bibitem[{Wei et~al.(2022{\natexlab{b}})Wei, Wang, Schuurmans, Bosma, Xia, Chi, Le, Zhou et~al.}]{wei2022chain}
Jason Wei, Xuezhi Wang, Dale Schuurmans, Maarten Bosma, Fei Xia, Ed~Chi, Quoc~V Le, Denny Zhou, et~al. 2022{\natexlab{b}}.
\newblock Chain-of-thought prompting elicits reasoning in large language models.
\newblock \emph{Advances in Neural Information Processing Systems}, 35:24824--24837.

\bibitem[{Ye and Durrett(2022)}]{ye2022unreliability}
Xi~Ye and Greg Durrett. 2022.
\newblock The unreliability of explanations in few-shot prompting for textual reasoning.
\newblock \emph{Advances in neural information processing systems}, 35:30378--30392.

\bibitem[{Zhao et~al.(2023)Zhao, Li, Joty, Qin, and Bing}]{zhao2023verify}
Ruochen Zhao, Xingxuan Li, Shafiq Joty, Chengwei Qin, and Lidong Bing. 2023.
\newblock Verify-and-edit: A knowledge-enhanced chain-of-thought framework.
\newblock \emph{arXiv preprint arXiv:2305.03268}.

\end{thebibliography}

\section{Language Resource References}
\label{lr:ref}
\bibliographystylelanguageresource{lrec-coling2024-natbib}
\bibliographylanguageresource{languageresource}

\section{Appendices}

\subsection{Human Annotation for Rationale Quality Measurement}
\label{sec:appendix_human_annotation}

\begin{figure*}[h]
\centering
\frame{\includegraphics[width=12cm]{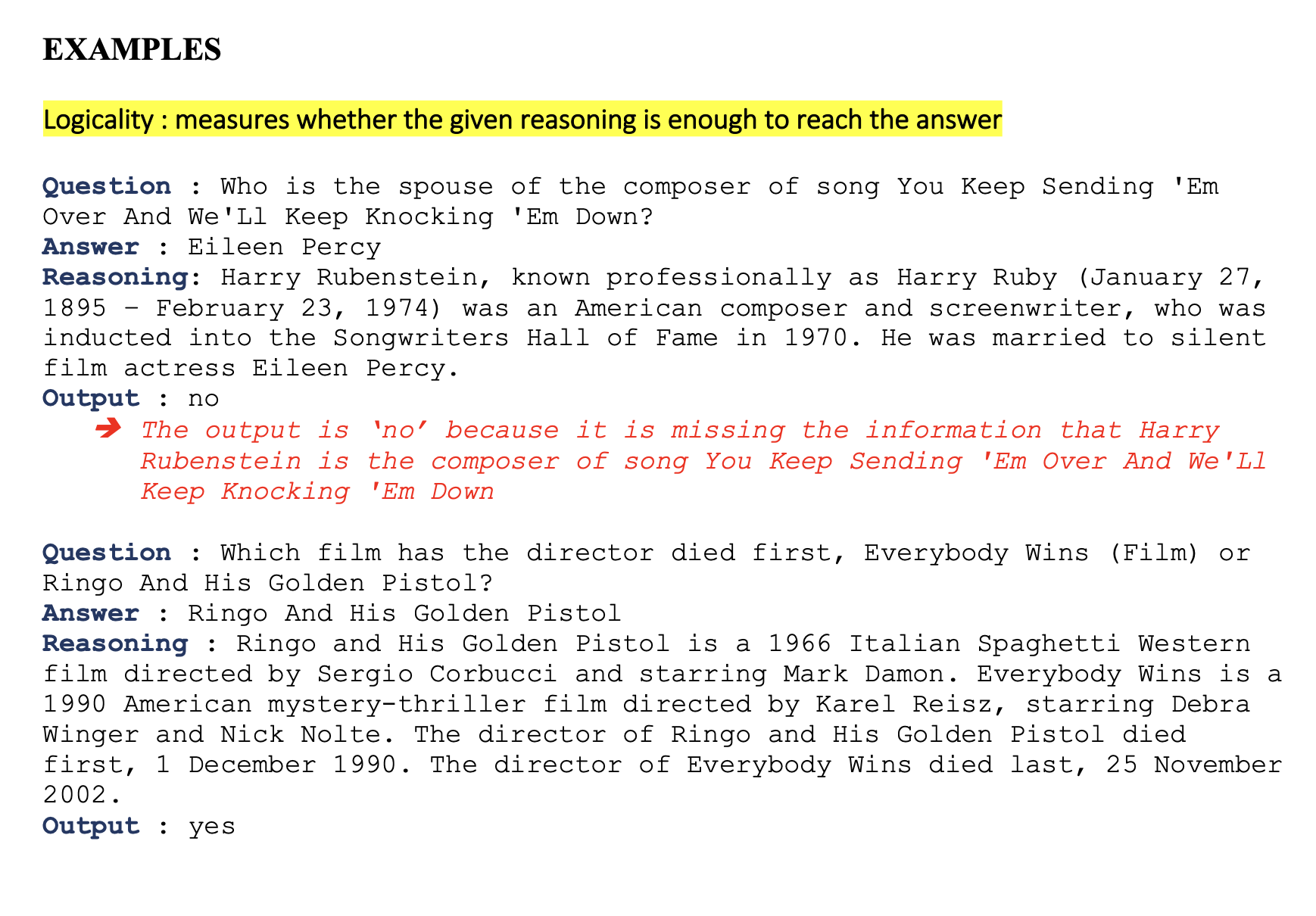}}
\caption{Demonstration example for "logicality" annotation.}
\label{fig:annotation_example}
\end{figure*}

\textbf{Annotation Details.} Three researchers manually inspected 100 instances randomly sampled from training data. To ensure the quality and consistency of the annotation process, one researcher designed the annotation instruction describing the definition of each aspect (see Table \ref{tab:metric_table}) as well as 4 demonstration examples for each aspect category. Figure \ref{fig:annotation_example} displays demonstration examples for logicality. These examples are chosen randomly from the remaining training set, which was not included in the initial set of 100 annotation samples, and are manually annotated based on aspect descriptions. Upon the completion of annotation, we gauged the inter-rater agreement rates to validate the reliability of submitted annotation results by computing the average Fleiss' Kappa coefficient \cite{fleiss1971measuring}. The mean Kappa score among the three annotators across all aspect categories was 0.56, indicating a moderate agreement rate. Ultimately, we take the mode of annotated labels submitted by three annotators and consider it as a ground truth. 

\begin{figure*}[h]
\centering
\includegraphics[width=13cm]{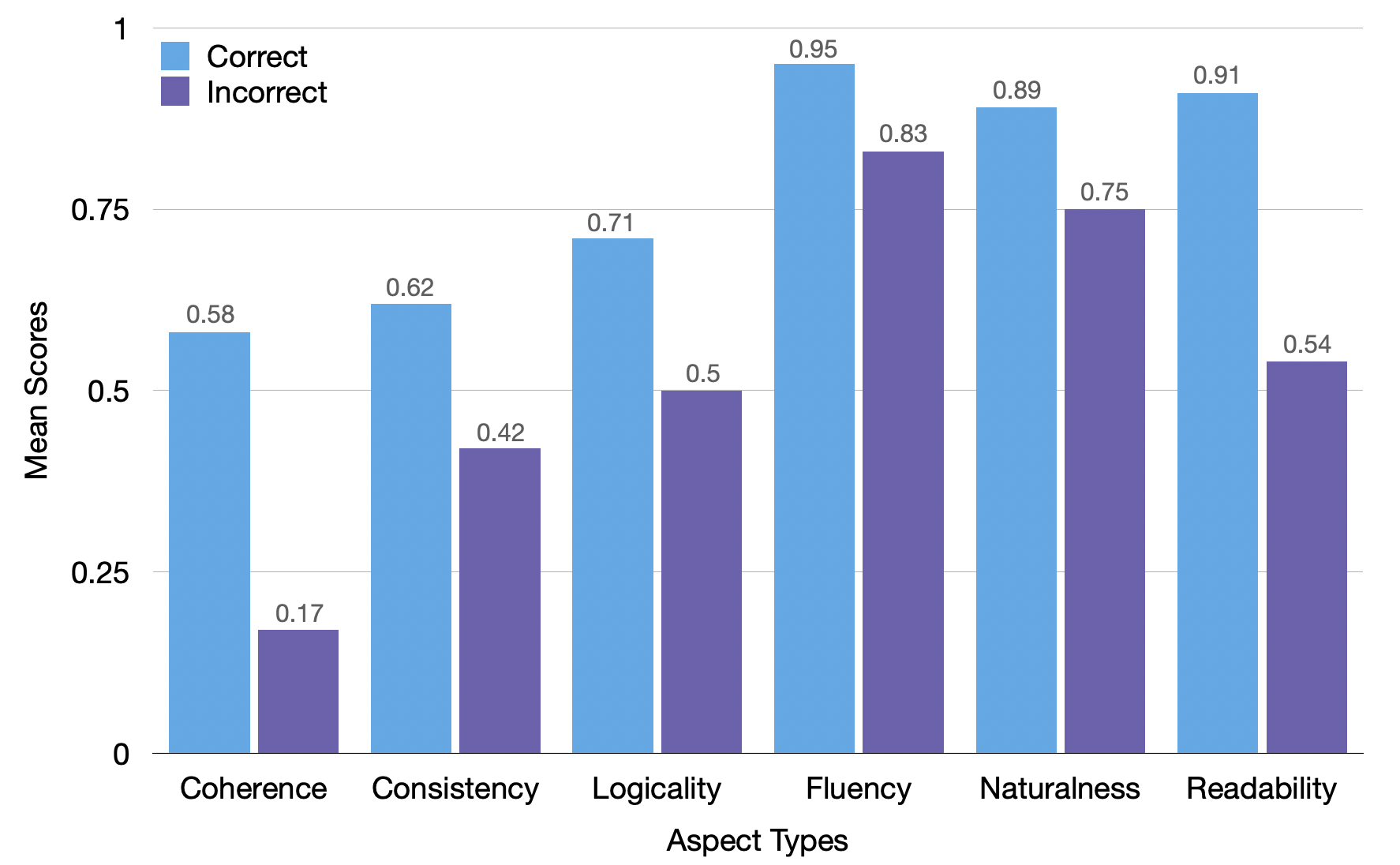}
\caption{Mean scores of human annotation results by answer prediction correctness.}
\label{fig:annotation_aspect}
\end{figure*}

\noindent \textbf{Analysis of Relationships between Aspect Types and Task Performance.}
Based on 100 labeled instances, we perform a post-hoc analysis to understand how the proposed aspects are related to answer prediction performance. The label distributions are as follows: correct (n=78) vs. incorrect (n=22). We compare the mean difference of evaluation scores based on correct \& incorrect responses. As shown in Figure \ref{fig:annotation_aspect}, rationales associated with correct answer prediction are prone to have higher scores than those with incorrect prediction. In particular, coherence has the largest mean difference, followed by readability and logicality. For fluency and naturalness, the gap seems minimal. We further conduct statistical testing to validate their statistical significance. T-test results confirm that the mean difference observed in coherence (\textit{p} = 0.003), readability (\textit{p} = 0.0002), and logicality (\textit{p} = 0.05) are statistically significant with \textit{p} < 0.05.

\subsection{Automatic Measurement for Rationale Quality}
\label{sec:appendix_automatic_annotation}

\begin{table*}[]
\small
\centering
\begin{tabular}{@{}cc|cccccc@{}}
\toprule
\multicolumn{2}{c|}{Prompting}                                     & Coherence     & Consistency   & Logicality    & Fluency       & Naturalness   & Readability   \\ \midrule
\multicolumn{1}{c|}{\multirow{2}{*}{IST}} & IST                    & 0.52          & \textbf{0.70} & 0.73          & 0.91          & \textbf{0.52} & \textbf{0.86} \\
\multicolumn{1}{c|}{}                     & IST + self-consistency & \textbf{0.53} & 0.67          & \textbf{0.74} & 0.90          & 0.50          & 0.85          \\ \midrule
\multicolumn{1}{c|}{\multirow{4}{*}{IDM}} & IDM (1 shot)           & \textbf{0.53} & 0.64          & 0.70          & \textbf{0.92} & 0.29          & 0.70          \\
\multicolumn{1}{c|}{}                     & IDM (2 shot)           & \textbf{0.53} & 0.62          & 0.73          & 0.91          & 0.24          & 0.69          \\
\multicolumn{1}{c|}{}                     & IDM (3 shot)           & 0.51          & 0.62          & 0.73          & 0.88          & 0.28          & 0.67          \\
\multicolumn{1}{c|}{}                     & IDM (4 shot)           & 0.52          & 0.62          & 0.70          & 0.89          & 0.36          & 0.50          \\ \bottomrule
\end{tabular}
\caption{The macro F1 scores for 5 prompt-based experiments from Method 1, based on 100 human-labeled examples. Values in bold represent the best performance in each aspect.}
\label{tab:method1_results}
\end{table*}

\begin{table*}[]
\small
\centering
\begin{tabular}{@{}c|ccccccccc@{}}
\toprule
\multirow{2}{*}{Methods} & \multicolumn{4}{c}{Coherence \& Consistency \& Logicality}   &  & \multicolumn{4}{c}{Fluency \& Naturalness \& Readability}   \\ \cmidrule(l){2-10} 
                         & Acc          & Precision     & Recall        & F1            &  & Acc          & Precision     & Recall        & F1           \\ \cmidrule(r){1-5} \cmidrule(l){7-10} 
Method 1                 & 0.62         & 0.62          & 0.6           & 0.6           &  & 0.7          & 0.7           & \textbf{0.81} & 0.67         \\
Method 2                 & \textbf{0.8} & \textbf{0.79} & \textbf{0.79} & \textbf{0.79} &  & \textbf{0.9} & \textbf{0.94} & 0.75          & \textbf{0.9} \\ \bottomrule
\end{tabular}
\caption{Evaluation results (Method 1 vs. Method 2). Values in bold represent the best performance in each aspect.}
\label{tab:comparison_results}
\end{table*}

\textbf{Method 1: Self-Evaluation from Large LMs.} \hypertarget{method1}{\citet{fu2023gptscore} have demonstrated the emergent capabilities of large LMs in neural text evaluation, achieved through zero-shot instruction and in-context learning. The key idea is that, given the natural language description of desired task and evaluation aspects, large LMs can assess multi-dimensional text quality without any learning process. Motivated by this, we instruct FLAN-T5 XXL to evaluate 6 aspects of the machine-generated rationales. Similar to \citet{fu2023gptscore}'s experiments, we investigate the performance of instruction-only (IST) prompting and instruction+demonstration (IDM) prompting. 
Let's say $q$ and $r'$ denote a question and a machine-generated rationale that is yet to be evaluated. $d$ represents the aspect definition of our interest from Table \ref{tab:metric_table}. A prompt template used for IST is:
\begin{mybox}
\footnotesize
Answer the question based on the provided information.
\newline Question: Can the given reasoning $d$ ? (a) Yes. (b) No.
\newline
\newline \textbf{Information}:
\newline Question: $q$
\newline Reasoning: $r'$
\newline Answer :
\end{mybox}
}
\noindent A prompt template for IDM is equivalent to IST but with the inclusion of a few task demonstrations. The number of demonstrations ranges from 1 to 4. We use the same demonstrations included in the human annotation instruction (\S \ref{sec:appendix_human_annotation}). Lastly, we evaluate IST in combination with SC decoding, which involves sampling the prediction multiple times (n = 10) and taking the mode as a final prediction. To ensure that the rationale evaluation of FLAN-T5 XXL aligns closely with human annotation, we compute the macro F1 scores for 5 prompt-based experiments using 100 human-labeled examples (Table \ref{tab:method1_results}). In most cases except for fluency, IST prompting demonstrates the highest performance.  

\noindent \textbf{Method 2: Supervised Training with Human-Annotated Data.}  
Here we train a logistic regression classifier using 100 ground truth data from \S \ref{sec:appendix_human_annotation}. This can be done by converting input data into TF-IDF vectors. Due to a small training data size, we resort to training two independent binary classifiers instead of having 6 models for each aspect type. While the first model is to predict if a given reasoning is logical, coherent, and consistent, the second model focuses on predicting whether the reasoning is fluent, natural, and readable. If at least one of the components is not satisfied, we consider it a negative label. The final label distribution is as follows: logicality \& consistency \& coherence (0: 60 vs. 1: 40), and fluency \& naturalness \& readability (0: 20 vs. 1: 80). We use 90\% of the dataset (n=90) for training and the remaining 10\% (n=10) for evaluation.

\noindent \textbf{Method 1 vs. Method 2.} We attempt to assess which one is more suitable for providing rewards for RL. For a fair comparison, we use the same evaluation data that was used in Method 2. Since Method 1 has 6 aspect categories, we obtain individual results using the best-performing approach and group them into two as we did for Method 2. Table \ref{tab:comparison_results} reports the accuracy, macro precision, macro recall, and macro F1 for Method 1 and Method 2. The results indicate that, although models from Method 2 are trained on a relatively small dataset, Method 2 is more aligned with human judgments compared to Method 1. Additionally, inference time using Method 2 is significantly faster than Method 1, making it easier to retrieve rewards in the RL setting. As a result, we use Method 2 for all our experiments. 

\subsection{Training Configuration}
\label{sec:appendix_training_config}
We provide a comprehensive description of the hyperparameters utilized in the model training process.

\noindent \textbf{Rationale Distillation.} 
For rationale generation from the teacher model, we randomly sampled 15K examples from each of the two datasets, resulting in a total of 30K examples. After filtering invalid rationales, the dataset was reduced to a total of 23K examples. 90\% of data was used for training, and the remaining 10\% was used for validation. For training, we used 8 NVIDIA Tesla V100 GPUs with 16GB configurations. Hyperparameters for training are as follows: 3e-3 for learning rates, 5 epochs, 64 batch size. 

\noindent \textbf{RL for Rationale Refinement.} For RL, we utilized the Huggingface's TRL\footnote{https://huggingface.co/docs/trl/index} library that provides a set of tools to train transformer LMs with RL. Instead of using the same examples used for the rationale distillation, we selectively chose 5000 examples that FLAN-T5 XXL failed to answer the question correctly. The reason behind this choice was to increase the model's exposure to challenging questions, thus increasing the likelihood of receiving more learning signals. We used 90\% of data for training and the remaining 10\% for validation. Training hyperparameters are as follows: 1.4e-5 for learning rates, 1 epoch, 16 batch size. For generation configurations, we set top\_k as 0.0, top\_p as 1.0, and enabled sampling.

\end{document}